\documentclass[a4paper]{jpconf} 

\usepackage{mathtools}
\usepackage{listings}
\usepackage{xcolor}
\usepackage{amssymb}
\usepackage{adjustbox}
\usepackage{float}
\usepackage{color,soul}
\usepackage{aliascnt}
\usepackage{amsmath}
\usepackage{textgreek}
\usepackage{caption}
\usepackage{tikz-cd}
\usepackage{graphicx}
\usepackage{multirow}
\usepackage{hyperref}
\usepackage{siunitx}
\restylefloat{table}

\begin{document}
\title{Neural Dynamic Movement Primitives - a survey}

\author{Jo\v{z}e M Ro\v{z}anec, Bojan Nemec}

\address{Jo\v{z}ef Stefan Institute, Jamova 39, 1000 Ljubljana, Slovenia}

\ead{\{joze.rozanec,bojan.nemec\}@ijs.si}


\begin{abstract}
One of the most important challenges in robotics is producing accurate trajectories and controlling their dynamic parameters so that the robots can perform different tasks. The ability to provide such motion control is closely related to how such movements are encoded. Advances on deep learning have had a strong repercussion in the development of novel approaches for Dynamic Movement Primitives. In this work, we survey scientific literature related to Neural Dynamic Movement Primitives, to complement existing surveys on Dynamic Movement Primitives.
\end{abstract}

\section{Introduction}
One of the most important challenges in robotics is producing accurate trajectories and controlling their dynamic parameters so that the robots can perform different tasks. The ability to provide such motion control is closely related to how such movements are encoded. The researchers explored different representations. First representations relied on spline curves \cite{chen1991solving}, which were used in tasks such as the recreation of arm movements \cite{lim2005movement}, or play Kendama \cite{miyamoto1996kendama}. While such representation allowed to reproduce certain movements, it lacked the flexibility to consider the time dimension \cite{schaal2007dynamics}, take into account perturbations \cite{wang2016dynamic}, and learn which control variables should be imitated and why \cite{ude2010task}. To mitigate such shortcomings, new approaches were developed, using Hidden Markov Models \cite{calinon2009statistical,osman2017trajectory}, Gaussian Mixture Models \cite{}, Neural Networks \cite{schaal2003control}, graph representations \cite{kallmann2004planning,manschitz2014learning}, and Dynamic Movement Primitives (DMPs) \cite{schaal2006dynamic,ijspeert2013dynamical}. 

The idea behind the Dynamic Movement Primitives was to develop a set of motion primitives that could be composed of more complex movements. They can be understood as a pattern generator, linear in its parameters, and invariant under rescaling \cite{peters2008reinforcement}. Each primitive is defined so that it can be adjusted without manual parameter tuning or introducing some instability. To that end, they have first defined a set of second-order differential equations to encode motion properties while taking into account possible perturbations \cite{schaal2006dynamic,ijspeert2013dynamical}. Furthermore, the equations provided for obstacle avoidance by encoding domain knowledge regarding working areas that must be avoided during the robot movements \cite{hoffmann2009biologically}, and provide means to execute the segments of such movements at different speeds \cite{schaal2007dynamics,ude2010task}. Multiple approaches were researched to determine the Dynamic Movement Primitives' parameters that determine the desired trajectory and task. One such approach is imitation learning, where an operator teaches the robot to perform a specific movement \cite{park2008movement,nemec2012action,meier2016probabilistic,bockmann2016kick}. The system can then generate a similar movement updating the scaling term to change the scope and duration of the movement \cite{yin2014learning}. Another approach is the use of Reinforcement Learning. While an initial set of parameters can be learned from initial demonstrations, Reinforcement Learning can be used to learn how changes in those parameters influence the movements \cite{kormushev2010robot}. Furthermore, Reinforcement Learning can be used to learn movement shapes without prior knowledge, ensuring the right policies are used. Among such methods we find the Natural Actor-Critic (NAC \cite{peters2008reinforcement}), the Policy Improvement with Path Integrals (PI\textsuperscript{2} \cite{theodorou2010reinforcement}), and Policy learning by Weighting Explorations with the Returns, (PoWER, \cite{kober2014policy}). The Dynamic Movement Primitives were successfully applied to encode periodic and discrete movements \cite{ijspeert2002movement,ijspeert2002learning}, in a wide variety of use cases, such as pick a glass of liquid \cite{nemec2012action}, kick a ball \cite{bockmann2016kick}, or perform some drumming \cite{ude2010task}.

While Dynamic Movement Primitives have shown value in representing complex movements, it has been noted that while they have been widely researched for trajectory-based skills, they were seldom applied to in-contact tasks \cite{hazara2016reinforcement}. Furthermore, it has been shown that the Dynamic Movement Primitives do not generalize well \cite{chen2015efficient}, and articulating multiple Dynamic Movement Primitives into complex motion sequences remains a challenge \cite{kormushev2010robot,li2017reinforcement}.

This research does not mean to provide an exhaustive survey on Dynamic Movement Primitives. For that purpose, we defer the reader to the survey on this topic by Saveriano et al. \cite{saveriano2021dynamic}. Nevertheless, we describe related work that attempts to solve the issues mentioned above by using neural networks and complement the survey regarding the use of neural networks and reinforcement learning for Dynamic Movement Primitives.

\section{Neural Dynamic Movement Primitives}
The first formulations of dynamic movement primitives were thought of as second-order differential equations to encode motion properties. Nevertheless, greater flexibility was introduced through the use of some learning components. \cite{zhang2021robot} developed a model that combined Gaussian Mixture Models and Gaussian Mixture Regression in order to learn a single Dynamic Movement Primitive from multiple demonstrations. A Neural Network controller to account for the influence of payloads on the movement dynamics and keep track of the reference trajectory modeled in joint space.
Furthermore, \cite{si2021composite} described the use of composite Dynamic Movement Primitives to learn manipulation skills while taking into account position and orientation constraints and the use of Neural Networks to approximate the forcing term describing the time-dependent process. While such representations have been helpful to represent movements, it has been also noted that the relationship between the Dynamic Movement Primitives' parameters and the configuration or task space is not evident, resulting in a lack of generalization on such spaces \cite{chen2015efficient}. Bitzer et al. \cite{bitzer2009latent} found that Dynamic Movement Primitives' parameters can directly relate to task variables when learned in latent spaces obtained when reducing the dimensionality of trajectories. This enables Dynamic Movement Primitives to better adapt to new task requirements, regardless of the task parameters. While in the aforementioned research, the dimensionality reduction was performed through a Gaussian Process Latent Variable Model, a similar approach was later attempted using autoencoders. Among the advantages of autoencoders, we find the ability to extract non-linear relationships between features and robustness to data, which does not follow a Gaussian distribution. Chen et al. \cite{chen2015efficient}, fitted the Dynamic Movement Primitives into an autoencoder (AE-DMP) and forced the model to learn movements in the feature space, avoiding hand-engineered features. They explored using two types of autoencoders: a basic autoencoder and a denoising autoencoder. While the basic autoencoder consists of an encoder and decoder networks, the denoising autoencoder corrupts the input during the training process and learns to successfully reconstruct the corrupted input into the expected output. They found that using data corresponding to multiple motions, the AE-DMP could switch motions and even generate motions not present among the training examples, based on existing ones. Using a sparse AE-DMP provides an additional advantage: hidden units are only rarely active, so their activation can be considered a source of explainability. A similar approach was followed by Chen et al. \cite{chen2016dynamic}, who used Deep Variational Bayes Filtering to embed DMPs into the latent space. Deep Variational Bayes Filtering learns a model with a fixed prior and simultaneously approximates an intractable posterior distribution with an approximate variational distribution, minimizing the Kullback–Leibler divergence. Furthermore, they simultaneously learn transitions to shape the latent space, backpropagating errors through time. Transitions over time and the approximate intractable posterior distribution are learned with a neural network, leveraging the fact that neural networks can work as universal function approximators \cite{HORNIK1989359}. The results show that the framework enables the representation of movements in a low-dimensional latent space, with excellent generalization properties and achieving optimal movements when reproducing the movements. While the authors mentioned above focused on learning movement transitions, \cite{dermy2018prediction} developed an approach to predict future intended movements. Such prediction is of utmost importance in use cases such as robot collaboration and the detection of unhealthy movements in assistive technologies. The number of observations can vary for each case for the task at hand. Given the prediction, computation time increases quadratically with the size of data used to represent the Probabilistic Movement Primitives (ProMP), dimensionality reduction techniques can reduce the ProMPs vector size and thus achieve better computation times. The authors demonstrated that such an approach strongly reduced the computational time required to issue a prediction, making such a method suitable for real-time applications.
While most of the aforementioned authors focused on representing Dynamic Movement Primitives in a latent space, Noseworthy et al. \cite{noseworthy2020task} focused on how such representations can be achieved by differentiating task and manner parameters. This direction of research was previously explored for classical Dynamic Movement Primitives by Matsubara et al. \cite{matsubara2010learning}. Noseworthy et al., on the other side, developed a Task-Conditioned Variational Autoencoder (TC-VAE), which conditions task parameters when learning from demonstrations, while an adversarial loss and an auxiliary network are used to train the encoder and ensure no information regarding the task parameters is encoded. By doing so, they achieved an interpretable, latent Dynamic Movement Primitive representation, which can generalize to task instances unseen during the training time. The conditioning of Neural Movement Primitives was also explored by Seker et al. \cite{seker2019conditional} who developed a framework for movement learning and generation (Conditional Neural Movement Primitives) built on top of Conditional Neural Processes. The framework sampled observations from the training set to learn conditional distributions regarding the target points. Then it learned temporal sensorimotor relations related to external parameters (e.g., sensor readings at each time step) and goals to generate trajectories and execute them gathering input from a feedback loop. The fact that sensor readings at each time step are considered, it enables the Conditional Neural Movement Primitives to detect unexpected events and change the movement trajectory as required. The applicability of Conditional Neural Movement Primitives was improved by \cite{Akbulut2020AdaptiveCN}, who coupled the Conditional Neural Movement Primitives with Reinforcement Learning to exploit formed latent representations and generate synthetic demonstrations that enable further learning. The Reinforcement Learning agent was triggered when a task was not properly executed so that a new trajectory could be learned from the most recent demonstration, leveraging shared characteristics with the original demonstrations. This way, the system learned to cope with new situations not contemplated in the original training set. The Conditional Neural Movement Primitives were also successfully applied to whole human body motion learning, where from a few motions, they were able to generate close to perfect trajectories of different body parts \cite{kurtoglu2020predicting}. Akbulut et al. \cite{akbulut2021acnmp} further developed the aforementioned work, who developed a framework named Adaptive Conditional Neural Movement Primitives to learn the distribution of input movement trajectories conditioned on the user set task parameters. The authors used an encoder-decoder network to represent the movement trajectories in a latent space. Such movements were learned from demonstrations and adapted to new trajectories or robots given new demonstrations, or Reinforcement Learning guided actions. A refined version of the framework was described in \cite{akbulut2021reward}, where a novel reward conditioned neural movement primitive was proposed to generate trajectories by sampling from the latent space conditioned on rewards. An encoder-decoder architecture was used to learn the distribution of trajectory points over time and the corresponding rewards to output a trajectory as a function of time.

While the authors mentioned above focused on finding means to learn Dynamic Movement Primitives from demonstrated trajectories, Pervez et al. \cite{pervez2017learning} designed task parameterized Dynamic Movement Primitives (TP-DMPs) which can adapt learned skills to different situations leveraging Convolutional Neural Networks (CNNs) to gather input from camera images. This architecture enables them to gather task parameters directly from such images and eventually generalize skills to arbitrary objects through data augmentation. The architecture has an input layer for the image and several layers to transform the image to feature task parameters. These are concatenated to a clock signal and passed to fully connected layers, which output the forcing terms. Following this same line of study, Gams et al. \cite{gams2018deep} proposed an image-to-motion encoder-decoder network to map raw images to Dynamic Movement Primitives. The architecture consisted of fully-connected layers only and was tested on the MNIST dataset\footnote{The MNIST dataset is available at the following URL: \url{http://yann.lecun.com/exdb/mnist/}}. The results showed that handwriting movements for digits could be reconstructed from raw images of digits. Ridge et al. \cite{ridge2020training} extended this work with two different architectures (IMED-Net (fully-connected), and CIMED-Net (convolutional)).

Among the central problems faced by researchers working on Deep Learning is data availability. Droniou et al. \cite{droniou2014learning} addressed this issue by using as little prior knowledge as possible to learn a set of actions. They do so through a deep learning architecture, which has an autoencoder to create a latent representation of a given point in time and use such a representation to learn a trace and the desired velocity. A memory layer then enables to execute movements at different velocities without explicitly modeling features. Another approach was explored by \cite{osa2019variational}, who developed the autoencoder trajectory primitives, encoding trajectory primitives, which can be used to generate synthetic trajectories to learn motion planning. To that end, first train a variational autoencoder with discrete and continuous latent variables, to then sample perturbations at the goal configuration following a Gauss distribution, and let the decoder generate a trajectory from a given goal position. To ensure constraints, such as joint limits, trajectory optimization methods are used to project the generated trajectory to the constrained solution space. Among the advantages of the solution, the authors describe the ability to use the decoder as a motion planner, whose input is the goal position and trajectory types, and the interpretability of the latent representation. Finally, Lon{\v{c}}arevi{\'c} et al. \cite{lonvcarevic2021generalization} explored the use of a fully-connected autoencoder network to reduce the dimensionality of a cartesian DMP space. To overcome the data scarcity and issues associated with data gathering (e.g., task executions on real robots can be error-prone), the authors gathered a small number of samples with a real robot, and they created synthetic samples with Gaussian process regression. Their results have shown that dimensionality reduction can be effectively learned from such data. Furthermore, this, in turn, was shown to be beneficial when applying Reinforcement Learning to learn a task, given that the lower dimensionality enabled a faster convergence.

\section{Dynamic Movement Primitives and Reinforcement Learning}
The previous sections described how different strategies were developed to encode Dynamic Movement Primitives. However, the fact that traditional Dynamic Movement Primitives, represented as nonlinear differential equations, are hard to generalize to new behaviors has spurred research to find new means to encode such information. Furthermore, many researchers have found that while basic movement information can be encoded in the Dynamic Movement Primitives, Reinforcement Learning can improve the quality of the Dynamic Movement Primitives, achieve greater generalization, and eventually learn new movements. Furthermore, Theodorou et al. \cite{theodorou2010generalized} considered the Dynamic Movement Primitives as a special case of parameterized policies and introduced the PI\textsuperscript{2} algorithm, working on continuous state-action spaces. Among the advantages of the algorithm, the authors list the fact that is numerically robust, and the ease to implement Reinforcement Learning algorithms based on trajectory roll-outs.

Robot programming by demonstration is one of the central topics of the field of robotics \cite{Billard2008}. Given the difference in kinematic capabilities between a human demonstrator and a humanoid robot, there is a need to adapt human motion trajectories to stable robot movements. \cite{kober2008reinforcement} addressed this issue by using Dynamic Movement Primitives to obtain a compact representation of a movement obtained from a first demonstration and then use Reinforcement Learning to improve the policy based on trial and error. They did so, generalizing the Reinforcement Learning policy from the immediate reward to the episodic reinforcement learning approach. They found that the state-dependent exploration enabled faster policy learning methods than the widely adopted Gaussian explorations with constant variance. This finding led them to devise a new expectation maximization-based policy learning algorithm called PoWER. Dynamic Movement Primitives specify three components: goal, shape, and temporal scaling. When used with Reinforcement Learning, only the weights for shaping the Dynamic Movement Primitives are usually learned, assuming the goal and temporal scaling are predefined. Nevertheless, there are many tasks where the goal position is not well known and can affect the outcome. Dynamic Movement Primitives shape and goal need to be learned simultaneously in such cases. To achieve a successful solution, Tamosiunaite et al. \cite{tamosiunaite2011learning} used different methods for the learning of the two components (NAC for goal learning, and PI\textsuperscript{2} for weight learning), to avoid mutual interference. NAC is a policy gradient method that transforms the gradient using the Fisher information matrix to obtain a natural gradient to compute the optimal reward. PI\textsuperscript{2}, on the other side, is derived from stochastic optimal control considers works by adding noise to the Dynamic Movement Primitives weights during an exploration phase, to later update the weights to minimize the cost. The authors found the approach successful and capable of ignoring incorrect feedback information so that they did not influence convergence towards a successful task resolution. Further research on how Dynamic Movement Primitives can be enhanced with Reinforcement Learning was performed by \cite{lundell2017generalizing}, who developed a global model to map task parameters to policy parameters. While they used the Dynamic Movement Primitives to encode the demonstrated movements, Reinforcement Learning was used to optimize the shape parameters and generalize the task to unseen situations. In their paper, the authors stress the importance of the noise structure and propose using correlated noise to balance learning speed and prevent safety hazards that could result from high noise variance. A different setting was explored by \cite{ogrinc2013motion}, who developed a system to convert information regarding human movements captured through a low-cost RGB-D camera into stable humanoid robot movements. While movements are not guaranteed to be optimal, they can be subsequently improved with model-free reinforcement learning and therefore achieve enhanced movement stability and accuracy. For this particular use case, the authors noted that different conditions must be considered, whether the movement corresponds to the lower or upper body part. When considering the lower body part, the movements must consider that at least one feet remain grounded when imitating a movement, while the upper body has no such constraints.

Another line of research was pursued by authors who envisioned the principle of \textit{divide and conquer} could be applied when designing the Reinforcement Learning components and therefore obtain better results by dividing the problem into smaller pieces. In this direction, Stulp et al. \cite{stulp2011hierarchical} used hierarchical Reinforcement Learning to reduce the search space of planning and control using tasks decomposition in order to ease complex task learning. They represented tasks through Dynamic Movement Primitives and applied Reinforcement Learning to Dynamic Movement Primitive sequences optimizing the movement primitive considering the cost of the following primitive in the sequence. PI\textsuperscript{2} was used to search the policy parameters space and simultaneously learn the shape and goal parameters of a motion sequence with no need for a model regarding the robot or the environment on which the task is performed. Furthermore, the authors found that such an approach led to lower overall costs than would be achieved by optimizing each Dynamic Movement Primitive separately over the motion sequence. A different approach was followed by Kim et al. \cite{kim2018learning}, who built a neural network to represent the differential equation and then learned to generalize the movements by using an actor-critic algorithm for deep reinforcement learning while applying a hierarchical strategy to decompose complex tasks and reduce the robot's search space in a sparse reward setting. To that end, two controllers are used: a meta-controller to learn policies over intrinsic (problem independent) goals and a sub-controller to learn policies over actions to achieve the given goals (problem-specific). E.g., the meta-controller would learn how to optimize movements given a sequence of tasks (e.g., tasks required to pick and place an object), while the sub-controller would learn how to best perform each of the sub-tasks (e.g., approach an object, grasp it, transport it to another location, and place it on the desired location). The authors reported implementing the hierarchical reinforcement learning component with the PI\textsuperscript{2} algorithm to learn the goal parameters only. In this same direction, but prescind from the hierarchical Reinforcement Learning, \cite{li2017reinforcement} explored the use of Dynamic Movement Primitives and Reinforcement Learning for humanoid robot movement, proposing a novel scheme for coupled Dynamic Movement Primitive-based motion sequence planning. They considered a two-level strategy, using (i) a neural dynamic optimization algorithm for redundancy resolution, (ii) Dynamic Movement Primitives to model and learn joint trajectories, and (iii) Reinforcement Learning to re-learn the proposed trajectories while including manipulation uncertainties. The initial spatial information and feedback are obtained through a machine vision component. The first step obtained the Dynamic Movement Primitive target points by considering the optimal solutions computed by the deep neural network and using the Dynamic Movement Primitives to model and learn the joint trajectories. Finally, the PI\textsuperscript{2} was used to learn the trajectory models considering external perturbations.

While new movements can be thought to the robot, they should be able to acquire new skills through imitation and self-improvement strategies. \cite{kormushev2010robot} proposed encoding the demonstrated skills into a compact form through a Dynamic Movement Primitive and then using the PoWER Reinforcement Learning policy to adapt and improve the encoded skill while learning the coupling along with different motor control variables. While a similar outcome could be realized using other Reinforcement Learning policies, the authors highlight two PoWER advantages when tackling the aforementioned problem: (i) there is no need to specify a learning rate, and (ii) together with importance sampling makes better use of previous agent's experience to determine new exploratory parameters.

The use of Dynamic Movement Primitives with Reinforcement Learning is frequently conditioned to specific use case scenarios, which determine meaningful aspects on how both should interact to achieve successful learning. We illustrate this with three use cases. First, \cite{hazara2016reinforcement} describes the modeling of in-contact tasks. Modeling in-contact skills is challenging since it requires considering not only the trajectory of the movement but control of the pose and force too since each action changes the environment and does not allow for the generalization of the task. In particular, the authors demonstrate their approach to the wood planning task. They report using a Dynamic Movement Primitive to encode the kinematic and force profiles that of a successful operation, and the PI\textsuperscript{2} Reinforcement Learning algorithm to update the force policy. Furthermore, they developed a framework to improve in-contact tasks by exploring the demonstrated force profile, arguing that it is not safe to start the exploration from a zero force profile for such tasks. The second use case refers to clothing assistance \cite{joshi2017robotic}. Similar to the setting described by \cite{hazara2016reinforcement}, the fact that the shape of the clothing article constantly changes, it affects the task settings. Therefore, the authors experimented using force information to detect whether the clothing article got stuck and concluded that the orientation of the end-effector must not be ignored to achieve a successful outcome. Finally, \cite{colome2018dimensionality} addressed a similar use case, teaching a robot to fold clothes. To that end, they characterized the robot moves through Dynamic Movement Primitives. The Dynamic Movement Primitive is then used to initialize a Reinforcement Learning process, leveraging a linear dimensionality reduction in between, to reduce the optimization problem's dimensionality and, therefore, the cost required to solve it. The results show that using dimensionality reduction not only fulfilled the purpose of decreasing the overall computational costs but also led to better learning curves.


\section{Benchmarks}
Benchmarks are a common approach to provide systematic comparisons in many research fields. This section describes four benchmarks developed to evaluate capabilities related to Dynamic Movement Primitives. When evaluating the Dynamic Movement Primitives, it is essential to consider how accurate the movements are when reproducing a demonstration and their robustness to perturbations and uncertainties.

The first such benchmark was proposed by Lemme et al. \cite{lemme2015open}, who described a framework to evaluate the performance of reaching motion generation approaches learning from a set of demonstrated examples. To that end, they provided a training set with human demonstrated trajectories (the LASA human handwriting library \footnote{The dataset is available at the following URL: \url{https://cs.stanford.edu/people/khansari/download.html}}), performed statistically sampled perturbations and evaluated the resulting movements across four scenarios and ten different metrics (at a geometric and/or kinematic levels). The metrics informed whether the motion reaches a specific target and the human-likeness of its trajectory. The benchmark considered only the motion generation of already trained models and therefore remains agnostic to the learning process. The authors aimed to evaluate the model's generalization to different initial conditions through the proposed scenarios of how discrete or continuous pushes or the target displacement affect the end-effector. Moberg et al. \cite{moberg2008benchmark} described a benchmark that evaluated the robustness of Dynamic Movement Primitives in the face of perturbations for a system consisting of a manipulator model (two-link manipulator with gear transmissions described by nonlinear friction and elasticity), and a discrete-time feedback controller. The introduced perturbations referred to measurement noise, a force, and torque disturbances applied at a particular execution segment. Their values are drawn from experience, regardless of the fact that they do not correspond to a specific robot design. Rana et al. \cite{rana2020benchmark} proposed a benchmark for motion-based learning from demonstration to evaluate how the complexity of the task, the expertise of the demonstrator, and the starting robot configuration affect the task performance. To that end, the authors considered four tasks (reaching, pushing, pressing, and writing) and nine participants with different experience levels (low, medium, high). They trained 180 task models and evaluated 720 task reproductions on a physical robot, measuring their mean squared error when contrasting the reproduced motions against the demonstrated trajectories. Finally, \cite{funk2021benchmarking} described a benchmark for the TriFinger opensource robotic platform, taking into account three policies for object manipulation and two data-driven optimization methods. The authors validated the policies across simulation and real robotic systems.

\section{Conclusion}
Dynamic Movement Primitives have shown to be an important abstraction to represent complex movements. Neural Networks as universal function approximators have enabled the robots to learn them from multiple inputs, such as images and demonstrations. Furthermore, it has enabled their representation at lower dimensional spaces while preserving nonlinearities.
The Dynamic Movement Primitives have been tested on various tasks and scenarios, showing great versatility. Finally, much work was invested into developing Dynamic Movement Primitives through the use of Reinforcement Learning or Dynamic Movement Primitives in the context of Reinforcement Learning. Dynamic Movement Primitives represented at low dimensions have proven crucial for achieving quick convergence of Reinforcement Learning in multiple settings. Given that benchmarks are key to providing a systematic comparison between methods, we concluded this work listing a set of benchmarks relevant to the evaluation of Dynamic Movement Primitives.


\section*{References}
\bibliographystyle{main}
\bibliography{main}

\providecommand{\newblock}{}
\begin{thebibliography}{10}
\expandafter\ifx\csname url\endcsname\relax
  \def\url#1{{\tt #1}}\fi
\expandafter\ifx\csname urlprefix\endcsname\relax\def\urlprefix{URL }\fi
\providecommand{\eprint}[2][]{\url{#2}}

\bibitem{chen1991solving}
Chen Y~C 1991 {\em Optimal Control Applications and Methods\/} {\bf 12}
  247--262

\bibitem{lim2005movement}
Lim B, Ra S and Park F~C 2005 Movement primitives, principal component
  analysis, and the efficient generation of natural motions {\em Proceedings of
  the 2005 IEEE international conference on robotics and automation\/} (IEEE)
  pp 4630--4635

\bibitem{miyamoto1996kendama}
Miyamoto H, Schaal S, Gandolfo F, Gomi H, Koike Y, Osu R, Nakano E, Wada Y and
  Kawato M 1996 {\em Neural networks\/} {\bf 9} 1281--1302

\bibitem{schaal2007dynamics}
Schaal S, Mohajerian P and Ijspeert A 2007 {\em Progress in brain research\/}
  {\bf 165} 425--445

\bibitem{wang2016dynamic}
Wang R, Wu Y, Chan W~L and Tee K~P 2016 Dynamic movement primitives plus: For
  enhanced reproduction quality and efficient trajectory modification using
  truncated kernels and local biases {\em 2016 IEEE/RSJ International
  Conference on Intelligent Robots and Systems (IROS)\/} (IEEE) pp 3765--3771

\bibitem{ude2010task}
Ude A, Gams A, Asfour T and Morimoto J 2010 {\em IEEE Transactions on
  Robotics\/} {\bf 26} 800--815

\bibitem{calinon2009statistical}
Calinon S and Billard A 2009 {\em Advanced Robotics\/} {\bf 23} 2059--2076

\bibitem{osman2017trajectory}
Osman A~A, El-Khoribi R~A, Shoman M~E and Shalaby M~W 2017 {\em Egyptian
  Informatics Journal\/} {\bf 18} 171--180

\bibitem{schaal2003control}
Schaal S, Peters J, Nakanishi J and Ijspeert A 2003 Control, planning,
  learning, and imitation with dynamic movement primitives {\em Workshop on
  Bilateral Paradigms on Humans and Humanoids: IEEE International Conference on
  Intelligent Robots and Systems (IROS 2003)\/} pp 1--21

\bibitem{kallmann2004planning}
Kallmann M, Bargmann R and Mataric M 2004 Planning the sequencing of movement
  primitives {\em proceedings of the international conference on simulation of
  adaptive behavior (SAB)\/} pp 193--200

\bibitem{manschitz2014learning}
Manschitz S, Kober J, Gienger M and Peters J 2014 Learning to sequence movement
  primitives from demonstrations {\em 2014 IEEE/RSJ International Conference on
  Intelligent Robots and Systems\/} (IEEE) pp 4414--4421

\bibitem{schaal2006dynamic}
Schaal S 2006 Dynamic movement primitives-a framework for motor control in
  humans and humanoid robotics {\em Adaptive motion of animals and machines\/}
  (Springer) pp 261--280

\bibitem{ijspeert2013dynamical}
Ijspeert A~J, Nakanishi J, Hoffmann H, Pastor P and Schaal S 2013 {\em Neural
  computation\/} {\bf 25} 328--373

\bibitem{peters2008reinforcement}
Peters J and Schaal S 2008 {\em Neural networks\/} {\bf 21} 682--697

\bibitem{hoffmann2009biologically}
Hoffmann H, Pastor P, Park D~H and Schaal S 2009 Biologically-inspired
  dynamical systems for movement generation: Automatic real-time goal
  adaptation and obstacle avoidance {\em 2009 IEEE International Conference on
  Robotics and Automation\/} (IEEE) pp 2587--2592

\bibitem{park2008movement}
Park D~H, Hoffmann H, Pastor P and Schaal S 2008 Movement reproduction and
  obstacle avoidance with dynamic movement primitives and potential fields {\em
  Humanoids 2008-8th IEEE-RAS International Conference on Humanoid Robots\/}
  (IEEE) pp 91--98

\bibitem{nemec2012action}
Nemec B and Ude A 2012 {\em Robotica\/} {\bf 30} 837--846

\bibitem{meier2016probabilistic}
Meier F and Schaal S 2016 {\em arXiv preprint arXiv:1612.05932\/}

\bibitem{bockmann2016kick}
B{\"o}ckmann A and Laue T 2016 Kick motions for the nao robot using dynamic
  movement primitives {\em Robot World Cup\/} (Springer) pp 33--44

\bibitem{yin2014learning}
Yin X and Chen Q 2014 Learning nonlinear dynamical system for movement
  primitives {\em 2014 IEEE International Conference on Systems, Man, and
  Cybernetics (SMC)\/} (IEEE) pp 3761--3766

\bibitem{kormushev2010robot}
Kormushev P, Calinon S and Caldwell D~G 2010 Robot motor skill coordination
  with em-based reinforcement learning {\em 2010 IEEE/RSJ international
  conference on intelligent robots and systems\/} (IEEE) pp 3232--3237

\bibitem{theodorou2010reinforcement}
Theodorou E, Buchli J and Schaal S 2010 Reinforcement learning of motor skills
  in high dimensions: A path integral approach {\em 2010 IEEE International
  Conference on Robotics and Automation\/} (IEEE) pp 2397--2403

\bibitem{kober2014policy}
Kober J and Peters J 2014 Policy search for motor primitives in robotics {\em
  Learning Motor Skills\/} (Springer) pp 83--117

\bibitem{ijspeert2002movement}
Ijspeert A~J, Nakanishi J and Schaal S 2002 Movement imitation with nonlinear
  dynamical systems in humanoid robots {\em Proceedings 2002 IEEE International
  Conference on Robotics and Automation (Cat. No. 02CH37292)\/} vol~2 (IEEE) pp
  1398--1403

\bibitem{ijspeert2002learning}
Ijspeert A~J, Nakanishi J and Schaal S 2002 Learning rhythmic movements by
  demonstration using nonlinear oscillators {\em Proceedings of the ieee/rsj
  int. conference on intelligent robots and systems (iros2002)\/} CONF pp
  958--963

\bibitem{hazara2016reinforcement}
Hazara M and Kyrki V 2016 Reinforcement learning for improving imitated
  in-contact skills {\em 2016 IEEE-RAS 16th International Conference on
  Humanoid Robots (Humanoids)\/} (IEEE) pp 194--201

\bibitem{chen2015efficient}
Chen N, Bayer J, Urban S and Van Der~Smagt P 2015 Efficient movement
  representation by embedding dynamic movement primitives in deep autoencoders
  {\em 2015 IEEE-RAS 15th international conference on humanoid robots
  (Humanoids)\/} (IEEE) pp 434--440

\bibitem{li2017reinforcement}
Li Z, Zhao T, Chen F, Hu Y, Su C~Y and Fukuda T 2017 {\em IEEE/ASME
  Transactions on Mechatronics\/} {\bf 23} 121--131

\bibitem{saveriano2021dynamic}
Saveriano M, Abu-Dakka F~J, Kramberger A and Peternel L 2021 {\em arXiv
  preprint arXiv:2102.03861\/}

\bibitem{zhang2021robot}
Zhang Y, Li M and Yang C 2021 {\em Neurocomputing\/} {\bf 451} 205--214

\bibitem{si2021composite}
Si W, Wang N and Yang C 2021 {\em Neural Computing and Applications\/}  1--11

\bibitem{bitzer2009latent}
Bitzer S and Vijayakumar S 2009 Latent spaces for dynamic movement primitives
  {\em 2009 9th IEEE-RAS International Conference on Humanoid Robots\/} (IEEE)
  pp 574--581

\bibitem{chen2016dynamic}
Chen N, Karl M and Van Der~Smagt P 2016 Dynamic movement primitives in latent
  space of time-dependent variational autoencoders {\em 2016 IEEE-RAS 16th
  international conference on humanoid robots (Humanoids)\/} (IEEE) pp 629--636

\bibitem{HORNIK1989359}
Hornik K, Stinchcombe M and White H 1989 {\em Neural Networks\/} {\bf 2}
  359--366 ISSN 0893-6080
  \urlprefix\url{https://www.sciencedirect.com/science/article/pii/0893608089900208}

\bibitem{dermy2018prediction}
Dermy O, Chaveroche M, Colas F, Charpillet F and Ivaldi S 2018 Prediction of
  human whole-body movements with ae-promps {\em 2018 IEEE-RAS 18th
  International Conference on Humanoid Robots (Humanoids)\/} (IEEE) pp 572--579

\bibitem{noseworthy2020task}
Noseworthy M, Paul R, Roy S, Park D and Roy N 2020 Task-conditioned variational
  autoencoders for learning movement primitives {\em Conference on robot
  learning\/} (PMLR) pp 933--944

\bibitem{matsubara2010learning}
Matsubara T, Hyon S~H and Morimoto J 2010 Learning stylistic dynamic movement
  primitives from multiple demonstrations {\em 2010 IEEE/RSJ International
  Conference on Intelligent Robots and Systems\/} (IEEE) pp 1277--1283

\bibitem{seker2019conditional}
Seker M~Y, Imre M, Piater J~H and Ugur E 2019 Conditional neural movement
  primitives. {\em Robotics: Science and Systems\/} vol~10

\bibitem{Akbulut2020AdaptiveCN}
Akbulut M~T, Seker M~Y, Tekden A~E, Nagai Y, {\"O}ztop E and Ugur E 2020 {\em
  ArXiv\/} {\bf abs/2003.11334}

\bibitem{kurtoglu2020predicting}
Kurtoglu M~H, Seker Y, Samur E and Ugur E 2020

\bibitem{akbulut2021acnmp}
Akbulut M, Oztop E, Seker M~Y, Hh X, Tekden A and Ugur E 2021 Acnmp: Skill
  transfer and task extrapolation through learning from demonstration and
  reinforcement learning via representation sharing {\em Conference on Robot
  Learning\/} (PMLR) pp 1896--1907

\bibitem{akbulut2021reward}
Akbulut M~T, Bozdogan U, Tekden A and Ugur E 2021 Reward conditioned neural
  movement primitives for population-based variational policy optimization {\em
  2021 IEEE International Conference on Robotics and Automation (ICRA)\/}
  (IEEE) pp 10808--10814

\bibitem{pervez2017learning}
Pervez A, Mao Y and Lee D 2017 Learning deep movement primitives using
  convolutional neural networks {\em 2017 IEEE-RAS 17th international
  conference on humanoid robotics (Humanoids)\/} (IEEE) pp 191--197

\bibitem{gams2018deep}
Gams A, Ude A, Morimoto J {\em et~al.\/} 2018 Deep encoder-decoder networks for
  mapping raw images to dynamic movement primitives {\em 2018 IEEE
  International Conference on Robotics and Automation (ICRA)\/} (IEEE) pp
  5863--5868

\bibitem{ridge2020training}
Ridge B, Gams A, Morimoto J, Ude A {\em et~al.\/} 2020 {\em Neural Networks\/}
  {\bf 127} 121--131

\bibitem{droniou2014learning}
Droniou A, Ivaldi S and Sigaud O 2014 Learning a repertoire of actions with
  deep neural networks {\em 4th International Conference on Development and
  Learning and on Epigenetic Robotics\/} (IEEE) pp 229--234

\bibitem{osa2019variational}
Osa T and Ikemoto S 2019

\bibitem{lonvcarevic2021generalization}
Lon{\v{c}}arevi{\'c} Z, Pahi{\v{c}} R, Ude A, Gams A {\em et~al.\/} 2021 {\em
  Applied Sciences\/} {\bf 11} 1013

\bibitem{theodorou2010generalized}
Theodorou E, Buchli J and Schaal S 2010 {\em The Journal of Machine Learning
  Research\/} {\bf 11} 3137--3181

\bibitem{Billard2008}
Billard A, Calinon S, Dillmann R and Schaal S 2008 {\em Robot Programming by
  Demonstration\/} (Berlin, Heidelberg: Springer Berlin Heidelberg) pp
  1371--1394 ISBN 978-3-540-30301-5
  \urlprefix\url{https://doi.org/10.1007/978-3-540-30301-5_60}

\bibitem{kober2008reinforcement}
Kober J 2008 {\em Reinforcement learning for motor primitives\/} Ph.D. thesis
  Universit{\"a}t Stuttgart Stuttgart, Germany

\bibitem{tamosiunaite2011learning}
Tamosiunaite M, Nemec B, Ude A and W{\"o}rg{\"o}tter F 2011 {\em Robotics and
  Autonomous Systems\/} {\bf 59} 910--922

\bibitem{lundell2017generalizing}
Lundell J, Hazara M and Kyrki V 2017 Generalizing movement primitives to new
  situations {\em Annual Conference Towards Autonomous Robotic Systems\/}
  (Springer) pp 16--31

\bibitem{ogrinc2013motion}
Ogrinc M, Gams A, Petri{\v{c}} T, Sugimoto N, Ude A, Morimoto J {\em et~al.\/}
  2013 Motion capture and reinforcement learning of dynamically stable humanoid
  movement primitives {\em 2013 IEEE International Conference on Robotics and
  Automation\/} (IEEE) pp 5284--5290

\bibitem{stulp2011hierarchical}
Stulp F and Schaal S 2011 Hierarchical reinforcement learning with movement
  primitives {\em 2011 11th IEEE-RAS International Conference on Humanoid
  Robots\/} (IEEE) pp 231--238

\bibitem{kim2018learning}
Kim W, Lee C and Kim H~J 2018 Learning and generalization of dynamic movement
  primitives by hierarchical deep reinforcement learning from demonstration
  {\em 2018 IEEE/RSJ International Conference on Intelligent Robots and Systems
  (IROS)\/} (IEEE) pp 3117--3123

\bibitem{joshi2017robotic}
Joshi R~P, Koganti N and Shibata T 2017 Robotic cloth manipulation for clothing
  assistance task using dynamic movement primitives {\em Proceedings of the
  Advances in Robotics\/} pp 1--6

\bibitem{colome2018dimensionality}
Colom{\'e} A and Torras C 2018 {\em IEEE Transactions on Robotics\/} {\bf 34}
  602--615

\bibitem{lemme2015open}
Lemme A, Meirovitch Y, Khansari-Zadeh M, Flash T, Billard A and Steil J~J 2015
  {\em Paladyn, Journal of Behavioral Robotics\/} {\bf 6}

\bibitem{moberg2008benchmark}
Moberg S, {\"O}hr J and Gunnarsson S 2008 {\em IFAC Proceedings Volumes\/} {\bf
  41} 1206--1211

\bibitem{rana2020benchmark}
Rana M~A, Chen D, Williams J, Chu V, Ahmadzadeh S~R and Chernova S 2020
  Benchmark for skill learning from demonstration: Impact of user experience,
  task complexity, and start configuration on performance {\em 2020 IEEE
  International Conference on Robotics and Automation (ICRA)\/} (IEEE) pp
  7561--7567

\bibitem{funk2021benchmarking}
Funk N, Schaff C, Madan R, Yoneda T, De~Jesus J~U, Watson J, Gordon E~K,
  Widmaier F, Bauer S, Srinivasa S~S {\em et~al.\/} 2021 {\em IEEE Robotics and
  Automation Letters\/} {\bf 7} 478--485

\end{thebibliography}

\end{document}